\documentclass[]{interact}

\usepackage{epstopdf}
\usepackage[caption=false]{subfig}

\usepackage[natbibapa,nodoi]{apacite}
\usepackage{xcolor}

\theoremstyle{plain}

\theoremstyle{definition}

\theoremstyle{remark}

\begin{document}


\title{MissMarple : A Novel Socio-inspired Feature-transfer Learning Deep Network for Image Splicing Detection}

\author{
\name{Angelina L. Gokhale\textsuperscript{a}\thanks{CONTACT Angelina L. Gokhale. Email: angelina@scit.edu}, Dhanya Pramod\textsuperscript{a}, Sudeep D. Thepade\textsuperscript{b}, and Ravi Kulkarni\textsuperscript{c}}
\affil{\textsuperscript{a}Symbiosis Centre for Information Technology (SCIT) , Symbiosis International (Deemed University) (SIU), Maharashtra, India; \textsuperscript{b}Pimpri Chinchwad College of Engineering (PCCOE), Savitribai Phule Pune University (SPPU), Maharashtra, India; 
\textsuperscript{c}Swami Vivekananda Yoga Anusandhana Samsthana (S-VYASA) (Deemed University), Karnataka, India}
}

\maketitle

\begin{abstract}
In this paper we propose a novel socio-inspired convolutional neural network (CNN) deep learning model for image splicing detection. Based on the premise that learning from the detection of coarsely spliced image regions can improve the detection of visually imperceptible finely spliced image forgeries, the proposed  model referred to as, \textit{MissMarple}, is a twin CNN network involving feature-transfer learning. Results obtained from training and testing the proposed model using the benchmark datasets like Columbia splicing, WildWeb, DSO1 and a proposed dataset titled AbhAS consisting of realistic splicing forgeries revealed improvement in detection accuracy over the existing deep learning models.
\end{abstract}


\section{Introduction}

Image manipulation is a growing concern especially with the advent of deep learning approaches such as generative adversarial networks presenting convincing realistic fakes (\citealp{goodfellow-gan}; \citealp*{dang2020}). In the past image forgery was performed using simple programming techniques to create image clones and composites occasionally rescaling the forged objects and post-processing the boundaries of these fake regions. Sophisticated tools like Photoshop made it even easier to perform these forgeries. Machine learning techniques specifically deep learning algorithms offer a variety of solutions to further enhance the creation of convincing image fakery in combination with existing tools at hand. The deliberate attempts of image forgery are done to mislead people in areas like media, courtroom trials, medicine, fashion industry, scientific research and social media \citep{farid2009-survey}. 

The most popular image forgery category includes that of copying (or cutting) a portion from one image (called the source image) and pasting it onto another portion in the second image (called the destination image). This is called image splicing or image composite or cut-paste forgery. A number of techniques involving pixel-level analysis (\citealp*{popescu2004}; J. Wang et al., \citeyear{wang10}, \citealp*{mookdarsanit15, Qur15}), investigating camera introduced artefacts \citep{johnson2006-lca, Pop05, mahdian2009, mayer2018-lca} studying format based signatures (\citealp*{farid2009-jpeg}; Ye, Q. Sun \& E.C. Chang, \citeyear{ye-sun-2007}; \citealp*{kee2011}), identifying inconsistencies in physical and geometrical properties of images (\citealp*{johnson2006, Joh07, Joh071, Kee10}; W. Fan, K. Wang, Cayre \& Xiong, \citeyear{Fan12}, \citealp*{Kee14}; B. Peng, W. Wang, Dong \& Tan, \citeyear{Pen17})  allows us to detect convincing image splicing forgeries \citep{farid2009-survey, farid2016, farid2019}. This earlier work focused on using digital signal processing techniques for extracting features from authentic and forged regions in order to detect the presence of forgery. Later, these features were inputted into machine learning classifiers to obtain better decision statistic in classifying spliced vs. authentic images (Hsu \& S.F. Chang, \citeyear{hsu2006}; T.T. Ng \& S.F. Chang, \citeyear{ng-chang-2004-model}; \citealp*{cozzolino2014, rao2017})

With the evolution of deep learning techniques, feature extraction was handled by deep learning models such as convolutional neural networks (CNN) and the extracted features were inputted to machine learning classifiers. The properties of CNNs being able to identify local features that humans may not be able to capture helped in strengthening the classification and localisation of image splices \citep*{rao2017, pomari2018, cozzolino2017, cozzolino2019}. Recent work in the area of deep learning has additionally proposed end-to-end frameworks for feature extraction and classification of image splicing forgery (\citealp*{bappy2017}; Luo, A. Peng, Zeng, Kang \& L. Liu, \citeyear{luo2019}; Yaqi Liu \& X. Zhao, \citeyear{liu2020}; \citealp*{ahmed2020}) The main concern when using end-to-end deep learning approaches is the requirement of a large training dataset for the model to learn effectively \citep{rao2017, Ben11, Coz18}. By transferring the learning from models trained on large datasets to models consisting smaller datasets, helps resolve the problem of effectively training the newer model with fewer samples. In this case, the pre-trained model addresses a different but related problem as the one proposed by the newer model (Pan \& Q. Yang, \citeyear{Pan10}; \citealp*{Tor10, Yos14}). Most of this recent work using deep learning for image splicing detection capitalises on the strengths of transfer learning (\citealp{rao2017, Coz18, Sin19}; Rao, Ni \& H. Zhao, \citeyear{rao2020}).

Considering the advantages of using deep learning techniques for image forgery detection, we present an end-to-end deep learning model for image splicing classification. The proposed model named \textit{MissMarple} is a socio-inspired twin CNN that uses feature-transfer learning to address the problem of image splicing detection. It is based on the premise that learning features from the detection of coarsely spliced image regions can improve the detection of visually imperceptible finely spliced image forgeries. The first half of this twin CNN has 4 convolutional layers and is trained using the coarsely spliced visually perceptible image splicing forged dataset. The second half of this twin with 4 convolutional layers is then trained using a realistic finely spliced dataset. Additionally, trained weights from the third convolutional layer of the first half are then concatenated with the third convolutional layer of the second network representing feature-transfer learning thus involving a total of 5 convolutional layers in the second half of the twin network.

The paper is organised as follows: Section \ref{section:related_work} discusses the related work on deep learning techniques for image splicing detection. Further, Section \ref{section:transfer_learning} explains the different transfer learning approaches and elaborates hose employed by existing deep learning models for splicing detection. The proposed model is explained in detail in Section \ref{section:proposed_model} with the experimental settings (Section \ref{section:experimental_settings}) and results obtained (Section \ref{section:results}). Finally, in Section \ref{section:discussion} we discuss the limitations and summarise the findings of this study concluding with proposed future work.

\section{Related Work}\label{section:related_work}
\subsection{Image Splicing Forgery Detection}
We review related techniques in the area of digital signal processing and machine learning for image splicing detection specifically tested on the Columbia uncompressed image splicing dataset (Hsu \& S.F. Chang, \citeyear{hsu2006}). Further, we also discuss deep learning approaches applied to image splicing forgery detection.

\subsubsection{Digital signal processing techniques and machine learning classifiers}
Shi, C. Chen and W. Chen \citeyearpar{shi-chen-2007} proposed a natural model for image splicing detection by first, applying a multi-size block discrete cosine transform (MBDCT) to a given test image and then, extracting statistical moments of characteristic functions along with Markov transition probabilities as features. Using a SVM classifier, they further classified images to be spliced or authentic. Zhang, Zhao and Su \citeyearpar{zhang-zhao-2009} used Markov process based features and DCT features to detect image splicing while Li, Jing and Li \citeyearpar{li-jing-2010} extracted two groups of features, namely, discrete wavelet transform (DWT) based moment features and Hilbert Huang Transform (HHT) features. These features were further passed through a SVM classifier to predict forged images.

He et al. \citeyearpar{he-sun-2011} applied the approximate run length method in combination with DWT while Moghaddasi, Jalab, Noor and Aghabozorgi \citeyearpar{moghaddasi2014} used a combination of run-length run-number (RLRN) and PCA image representation features to detect image splicing. Muhammad, Al-Hammadi, Hussain and Bebis \citeyearpar{muhammad2014} demonstrated the use of steerable pyramid transform and local binary patterns (LBP) for detecting image splicing manipulation. Goh and Thing \citeyearpar{goh2015} used hybrid and ensemble models while Hussain, Muhammad, Saleh, Mirza and Bebis \citeyearpar{hussain2013} applied the Weber local descriptors. C. Li, Ma, Xiao, M. Li and A. Zhang \citeyearpar{li2017-qdct} applied quaternion features of DCT (QDCT) and Markov features while Z. He, W. Lu, W. Sun and Huang \citeyearpar{he2012} combined Markov features with DCT and DWT transforms. On the other hand, Alahmadi et al. \citeyearpar{alahmadi2017} combined DCT coefficients with LBP to classify spliced images. J.G. Han, Park, Moon and Eom \citeyearpar{han2016} applied Markov features and expectation-maximization (EM) technique while Q. Zhang, W. Lu and Weng \citeyearpar{zhang-lu-2016} combined DCT and counterlet. B. Chen, Qi, X. Sun, Shi \citeyearpar{chen-qi-2017} used quaternion pseudo-Zernike moments combining both of RGB information and depth information. They empirically analysed variants of their proposed technique on the RGB and YCbCr space and achieved the best results for the YCbCr space. For seminal surveys on image forgery detection techniques, refer to (\citealp*{birajdar2013}; \citealp*{zampoglou2017}; Zheng, Y. Zhang and Thing, \citeyear{zheng2019} and \citealp*{farid2019})

\subsubsection{Deep learning techniques}
Deep learning (DL) algorithms have evolved in the last few years and have become a standard approach to solving image classification problems. These methods can be defined as those that learn through multiple layers of representations where at each layer, they search for meaningful representations of some input data, within a pre-defined space of possibilities with guidance from an input signal \citep*{chollet2016, good2016}. Despite the relative success of the former methods involving DSP and ML, the problem of effectively detecting traces of image manipulation prevails. The issue could partly be due to the inability to capture appropriate manipulation telltales. Hence, using the DL approach, learning the said mappings and descriptive features from the available training data becomes easier \citep{pomari2018}.

Successful attempts of applying DL techniques for detecting universal traces of image manipulation flourished since the year 2016.  Among some of the classification techniques, Bayar and Stamm \citep{bayar2016} devised a CNN-based universal forgery detection technique designed to suppress image content and adaptively learn manipulation detection features. The proposed work used a constrained CNN with 12 prediction error filters. This was later followed by two non-constrained CNNs that were able to detect image enhancement operations such as median filtering, Gaussian blurring, additive white Gaussian noise (ADGWN) and resampling. Rao and Ni \citeyearpar{rao2017} utilised a CNN to automatically learn hierarchical representations from RGB color images. The first layer of the CNN was constrained to suppress image content by initialising its weights to spatially rich models (SRM) residual maps \citep{fridrich2012}. Later, the dense features extracted by the pre-trained CNN were passed to the feature fusion technique followed by SVM classification. Recently, Rao, Ni and H. Zhao \citeyearpar{rao2020} extended their work to image splicing detection and localisation scheme by developing a two-branch CNN model that automatically learns hierarchical representations from the input RGB or monochrome images. The model is a constrained CNN like their previous work but they additionally  used contrastive and cross entropy loss in combination for improving the generalisabiltiy of their proposed model. Here, the final discriminative features were obtained by employing a block-pooling feature-fusion strategy which were then passed to an SVM classifier. The image splicing was localised by using the fully connected conditional random field (CRF) scheme. The model was also found to be robust against JPEG compression.

Mayer, Bayar and Stamm \citeyearpar{mayer2018-unified} designed unified deep-features for multiple forensic tasks. They used a transfer-learning approach where features learned for the camera detection tasks were then applied for detecting image manipulations. They noted task symmetry where the reverse was not true. Pomari et al. \citeyearpar{pomari2018} detected photographic splicing by combining the high representation power of illuminant maps and CNN to learn evidences of tampering directly from the training data, thus eliminating the laborious feature engineering process. The model was built on the ResNet-50 (K. He, Xiangyu Zhang, S. Ren \& J. Sun, \citeyear{HeK16}) architecture by optimizing only the final layers with an SVM classifier. The transfer learning CNN model nicknamed as deep splicing feature (DSF) extractor, learned  features from the illuminant maps \citep*{carvalho-2016} of pristine and spliced images. Finally, they applied the Grad-CAM method \citep{selvaraju2017} to localise splicing. Though they present results of their localisation approach, the experimental details for comparison are based on their classification accuracy.

During earlier work, a number of machine learning classifiers proved beneficial to separate the spliced images from the pristine ones. Even then, it became important to localise the region of tampering since it was crucial to understand which part of the image has been tampered. Important decisions were based on the outcome of the image authenticity test; therefore, recent work of forgery detection also concentrates on localising its traces. Cozzolino and Verdoliva \citeyearpar{cozzolino2016-autoencoder} tested the capability of auto-encoders to localise splicing by extracting noise residual features. They used a simple feedforward autoencoder with a single hidden layer that inputs SRM features \citep{fridrich2012} and outputs a discriminative label as 0 (pristine) or 1 (spliced) using Otsu's strategy \citep{otsu1979}.  Yanfei Liu, Zhong and Qin \citeyearpar{liu-zhong-2018} used CNN to extract statistical features which were earlier used by G. Xu, Wu and Y.Q. Shi \citeyearpar{xu-wu-2016} for their work on steganalysis. They proposed a multi-scale CNN (MSCNN) which inputs sliding windows of different scales and outputs real-valued tampering possibility maps. Further, using the simple linear iterative clustering (SLIC) \citep{achanta2012} method for constructing a graph on superpixels, the final decision map was created by fusing the obtained possibility maps which localised the tampered region. Y. Zhang, Goh, Win and Thing \citeyearpar{zhang-goh-2016} used a stacked auto-encoder which was trained in two-steps to learn complex features from input images and then applied context learning for localisation.

Bappy et al. \citeyearpar{bappy2017} combined long short-term memory (LSTM) with CNN to detect traces of image splicing. Bondi et al. \citep{bondi2017} used CNN based extraction of camera model characteristics from image patches followed by an iterative clustering analysis to detect and localise the tampered region. Cozzolino, Poggi and Verdoliva \citeyearpar{cozzolino2017} used spatially rich model (SRM) image features based on noise residuals earlier applied for steganalysis \citep{fridrich2012} to effectively localise splicing. Here, they tested the effect of first constraining earlier layers of CNN and then relaxing them to detect image manipulation. Salloum, Y. Ren and Kuo \citeyearpar{salloum2018} proposed a multi-task fully convolutional neural network (MFCN) which utilised two output branches for learning surface label and edges of the spliced region. Their edge-enhanced MFCN has proven the most effective in localising traces of splicing on CASIA v1 (Dong, W. Wang, \& Tan, \citeyear{dong2013-casia}), Columbia-color (Hsu \& S.F. Chang,\citeyear{hsu2006}) and the NIST-Multimedia Forensic Challenge 2018 datasets\footnote{https://www.nist.gov/itl/iad/mig/media-forensics-challenge-2018}. Huh, A. Liu, Owens and Efros \citeyearpar{huh2018} used Siamese Networks and self-consistency maps extracted from the EXIF (exchangeable image file format) image headers.

Yaqi Liu, X. Zhao, Zhu and Cao \citeyearpar{liu-zhao-2018} proposed a novel adversarial learning framework to train a constrained image splicing detection and localisation deep matching network.  Their framework consisted of 3 building blocks: the deep matching network based on atrous convolution (DMAC) which outputs two high-quality candidate masks indicating the suspected regions of the two input images; the detection network which rectifies the inconsistencies between the corresponding candidate masks and the discriminative network that drives the DMAC network to produce masks which are hard to distinguish from the ground-truth masks. In steps 2 and 3, they applied a variant of adversarial learning inspired from the work on generative adversarial networks \citep{goodfellow-gan} which they nicknamed as DMAC-adv-det. Their model showed very high localisation accuracy for the CASIA v2 dataset. Wei, Bi and Xiao \citeyearpar{wei-bi-2018} proposed a two CNN model, first, a coarse CNN (C-CNN) that learns from the coarse suspicious tampered regions revealing the edge of spliced regions; second, a refined CNN (R-CNN) that learns the property differences between the authentic side of the spliced edge and the spliced region itself. They initially used a patch-level approach for training their model and further developed an equivalent model that accepted image-level input. The image-level model did not hamper the performance. For C-CNN, the patches were selected by centering a pixel on the spliced edge boundary and then extracting a 32x32 window around that pixel. They further extended their work by introducing the diluted adaptive clustering approach \citep{xiao2020}. Their proposed method achieved improvement in image splicing detection.

Zhou, X. Han, Morariu and Davis \citeyearpar{zhou2018}, designed a two-stream faster Residual-CNN where first part of the model focused on the RGB stream for visual clues and second part on the noise distributions. Cozzolino and Verdoliva \citeyearpar{cozzolino2019} built a network based on the Siamese network concept, with two identical CNNs that learn noise residuals thus improving its extraction process. They enhanced the camera model artifacts for their direct use in forensic analyses. The output was an image-size, noise residual (called noiseprint) that bore traces of camera model artifacts thus helping in localisation. While comparing their method with the benchmarking list given by \citep{zampoglou2017} they added another algorithm to this list and referred it as NOI4 \citep{wagner-2015}. Luo et al. \citeyearpar{luo2019} proposed a deep-residual learning model for median filtering detection. They employed a data augmentation technique to overcome the problem of model over-fitting. Refer to \citep*{nadeem2019} a recent survey on deep learning techniques for multimedia forensics including splicing detection.

In their recent work, Abd El-Latif, Taha and Zayed \citeyearpar{latif2020} proposed a CNN model for extracting features from a spliced image. A discrete wavelet transform (DWT) was then applied to the output feature vector of this CNN model and the features were then passed on to a SVM classifier. They also compared their work by replacing the DWT process with discrete cosine transform (DCT) and principal component analysis (PCA) being applied to the output feature vector of the CNN model. They were able to achieve a high classification accuracy on publicly available image splicing datasets. Hussein, Mahmoud and Zayed \citeyearpar{hussien2020} extracted features from the authentic and spliced input images by analysing the color filter array (CFA) pattern. The features were then reduced by applying PCA and then passed through the deep belief network-deep neural network (DBN-DNN) classifier consisting of input layer3, output layer and six hidden layers.

Yaqi Liu and X. Zhao \citeyearpar{liu2020} proposed a novel deep matching network namely, AttentionDM, for constrained image splicing detection and localisation. It consists of an encoder-decoder model with atrous convolution for hierarchical features dense matching and fine-grained mask generation for localisation. They employed pre-trained models VGG16, ResNet50 and ResNet101 in their encoder-decoder architecture for feature extraction and integrated it with the atrous convolution module. Cozzolino et al. \citeyearpar{Coz18} and Dang et al. \citeyearpar{dang2020} developed deep learning models to detect tampered and computer-generated facial image forgeries. Refer to \citep*{Gok20} for an elaborate mind map overview of the different techniques involved in image forgery detection.

From the above literature, we see that a lot of recent work emphasises tamper localisation. We argue that though this is important, but with a research objective such as real-time stripping off unwarranted content from social networking websites, classification will supersede localisation. Section \ref{section:localization} briefly discusses the proposed localisation technique following classification of spliced images but rest of the discussion focuses on accurately classifying spliced images. 

\subsection{Transfer Learning}\label{section:transfer_learning}
Transfer learning is the process of transferring the learning (or knowledge) between different but related tasks \citep{Pan10, Tor10, Ben11}. Following the notations in \cite{Pan10}, we consider a source domain $D_S$ and a target domain $D_T$. The source and target tasks are referred to as $T_S$ and $T_T$ respectively. In case of transfer learning, we assume that $D_S \neq D_T$ and $T_S \neq T_T$. The learning is then transferred from $D_S$ to $D_T$. In case of deep learning, there is difficulty in obtaining large training sets for different problems. Transfer learning is a popular technique adopted in deep learning where knowledge from a model pre-trained on larger datasets is transferred to another model with different but related task at hand usually having insufficient number of training samples \citep{Ben11}.

Related work in deep-transfer learning includes (\citealp{Sin19, rao2017, luo2019}; Yaqi Liu et al., \citeyear{liu-zhao-2018}). There are various settings for transfer learning namely, inductive, transductive and unsupervised learning. In the case of transductive transfer learning, the target labels are not available while for unsupervised transfer learning both the source and target labels are missing \citep{Pan10}. In this paper we focus on inductive transfer learning where the source labels and target labels are present. In the case where the source labels are absent or not relevant it is self-taught transfer learning. Usually in inductive learning, $D_S = D_T$ while $T_S \neq T_T$ \citep{Pan10}. Further, there are different approaches to transfer learning namely, instance-transfer, feature-representation-transfer (or feature-transfer), parameter-transfer and relational-knowledge-transfer. For details on the transfer learning approaches and their relation to the different sub-settings, refer to \citep{Pan10}. Aditionally, to address the problem of image splicing detection, deep learning models used for the feature engineering process, also consider features extracted from models trained for a specific but related problem of interest. Rao and Ni \citeyearpar{rao2017} demonstrated a feature-transfer learning approach by using features from the residual maps of spatially rich models (SRM) previously used by Fridrich and Kodovsky \citeyearpar{fridrich2012} for steganalysis.

Our proposed model \textit{MissMarple}, demonstrates a feature-transfer learning approach under the inductive transfer learning setting. Here, $D_S = D_T$ and $T_S = T_T$. Though $D_S = D_T$, they are similar but not same, that is, the data in the respective domains is different. In feature-transfer learning approach, a "good" feature representation from $D_S$ that can reduce 	its difference from $D_T$ is considered and applied to $D_T$ \citep{Pan10}. We  discuss the implementation of feature-transfer learning for \textit{MissMarple} model in section \ref{section:design}.

\section{Proposed Model}\label{section:proposed_model}
\subsection{Motivation}\label{section:motivation}
Earlier research work based on machine learning approaches to image splicing detection showed an improved detection accuracy with realistic (visually imperceptible) image splices as compared to the accuracy achieved on the datasets with coarse image splices (visually perceptible) \citep{zheng2019}. For example, in (Park, J.G. Han, Moon and Eom, \citeyear{park2016}; \citealp{han2016}) results obtained on the Columbia image splicing dataset (Hsu \& S.F. Chang, \citeyear{hsu2006}) were not higher compared to the realistic CASIA v2. \citep{dong2013-casia}. Logically it seems natural to think that machine learning models should effectively detect visually perceptible forgeries but empirically this seemed to fail \citep{zheng2019}. Digital signal processing techniques applied alone or in combination with ML classifiers performed better in comparison with end-to-end ML techniques on the coarsely spliced Columbia dataset (see: Table \ref{table:table7}). Deep learning models too on the other hand showed better detection accuracy with realistic datasets like DSO-1 and DS-1 compared to coarsely spliced Columbia dataset \citep{pomari2018}. The design for the proposed model built on this premise. In order to strike a balance of the training datasets and improve detection accuracy on both coarsely spliced as well as finely spliced datasets, we decided to choose a combination of both these datasets for training. Table \ref{table:table2} discusses details of the datasets used for the proposed design.

\subsection{Intuition and Philosophy}
\begin{figure}
	\centering
	\includegraphics[width=11cm, height=11cm, keepaspectratio]{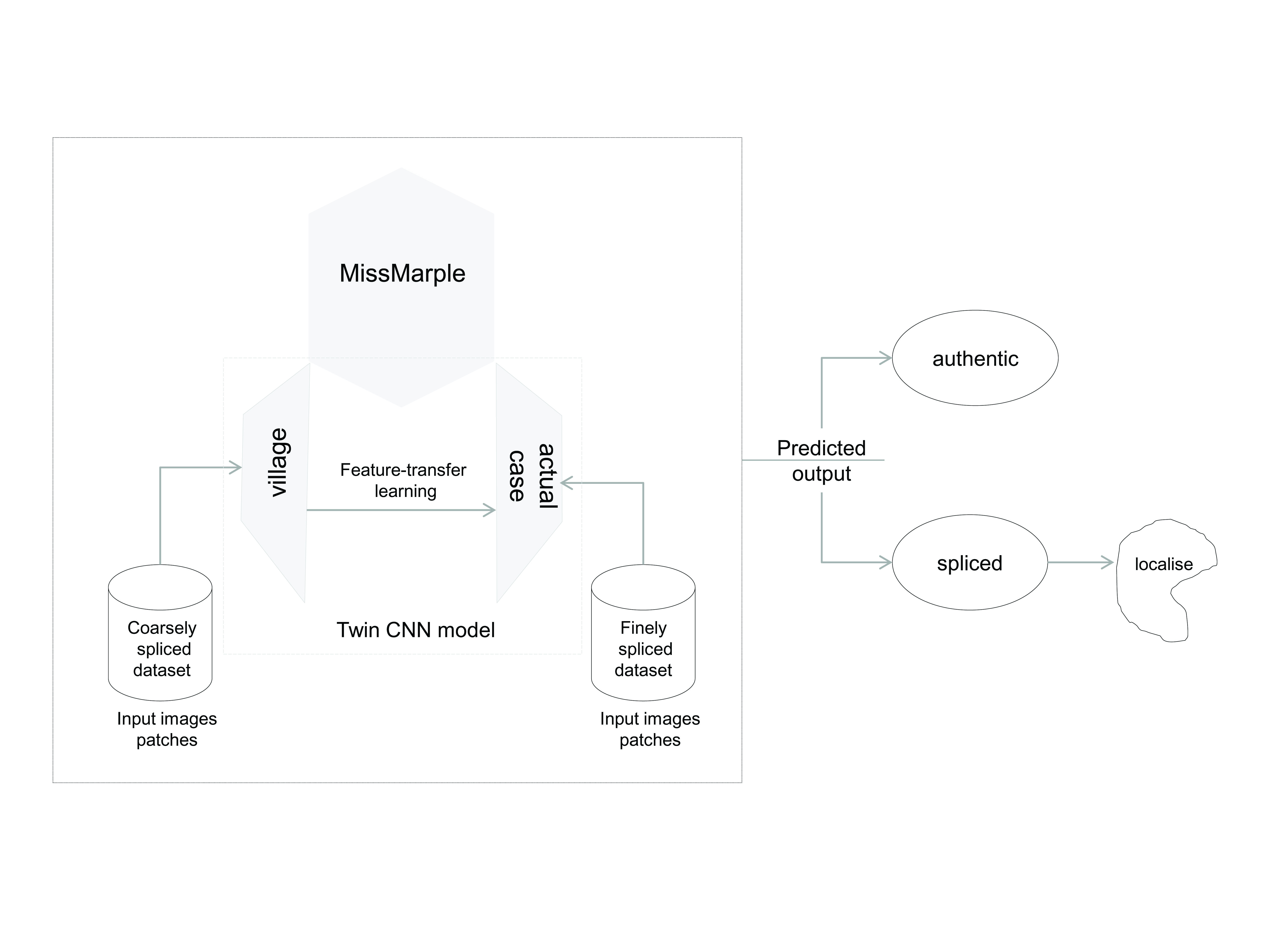}
	\caption{Overview of the proposed MissMarple design}\label{fig:fig1}
\end{figure}

Studies in developmental psychology and social psychology have often quoted Agatha Christie's fictional detective, Miss Jane Marple who employs psychological principles of human behaviour to solve crimes \citep{wesseling2004}. Her study of human behaviour in society (drawing inspiration from her village parallels)which she develops by observing people in her village helps her to solve crimes in real life \citep{beveridge1998}. The proposed model, \textit{MissMarple}, is therefore inspired from this fictional detectives ability to solve crimes. Based on the assumption that people in villages would lead a more simpler life compared to the sophistication of cities, the proposed model, designed as a twin convolutional neural network, is trained on two separate datasets namely, the coarse splicing Columbia dataset (Hsu \& S.F. Chang, \citeyear{hsu2006}) and the realistic finely spliced Wild Web dataset \citep{zampoglou2015}. The association drawn between the personality traits of villagers vs. the suspects of the crime, is realised through \textit{feature-transfer} learning between the twin convolutional neural networks. Figure \ref{fig:fig1} presents an overview of the overall proposed model design.

\subsection{Design} \label{section:design}
The socio-inspired proposed model \textit{MissMarple}, is a twin convolutional neural network (CNN). The first part of this model, named the \textit{village\_model} (MM-V) consists of 4 convolutional layers with alternating maxpooling layers. The primary design of this model is inspired from a basic CNN model for the dogs vs. cats classification \citep[pg.~130]{chollet2016}, which consists of a similar implementation of 4 convolutional layers. The maxpooling layers help in reducing the feature set allowing the model to generalise better. To avoid overfitting of this model, we add the BatchNormalisation layer alongwith the Dropout layer which also regularises the model. Finally, we add two fully connected dense layers with the \textit{sigmoid} activation for this binary classification to distinguish spliced images from the authentic ones. For all the convolutional layers, we use the \textit{relu} activation. Instead of processing the entire images of varied sizes and for overcoming the limitation of fewer training images, we pass fake patches of size 64x64 carefully selected by overlaying the masks on the respective fake images (with 40\% overlap) and similarly authentic patches of size 64x64 randomly selected from the authentic images. Thus, the dataset of fake vs. authentic patches is balanced. We adopt this ideology from \citep{Sin19} for extracting the 64x64 patches.

\begin{figure}
	\centering
	\includegraphics[width=10cm, keepaspectratio]{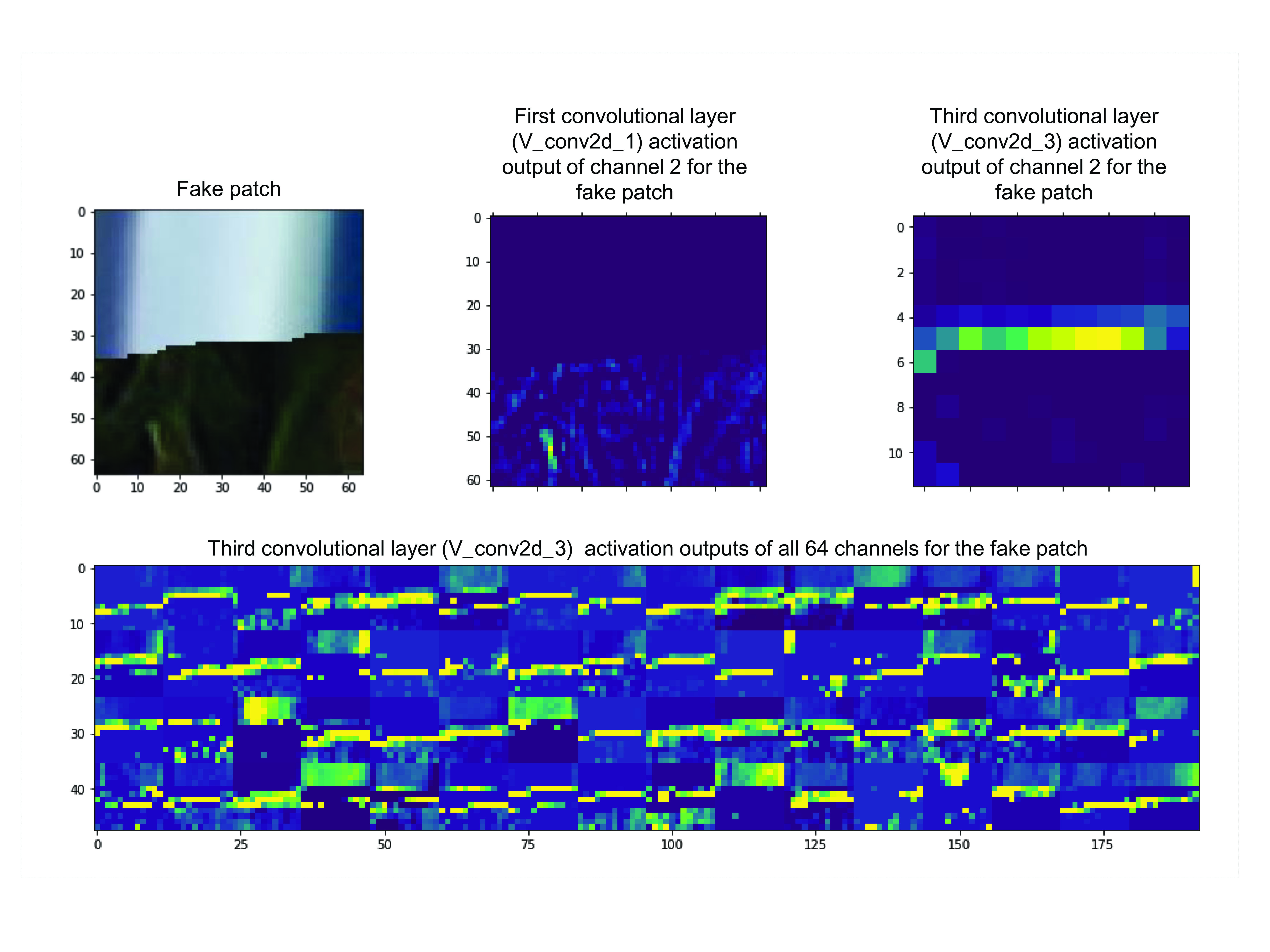}
	\caption{Visualising the output activations from the third convolutional layer (V\_conv2d\_3) of the proposed \textit{MissMarple} village (MM-V) model}
	\label{fig:fig2}
\end{figure}

\begin{figure}
	\centering
	\includegraphics[width=18cm, height=18cm, keepaspectratio]{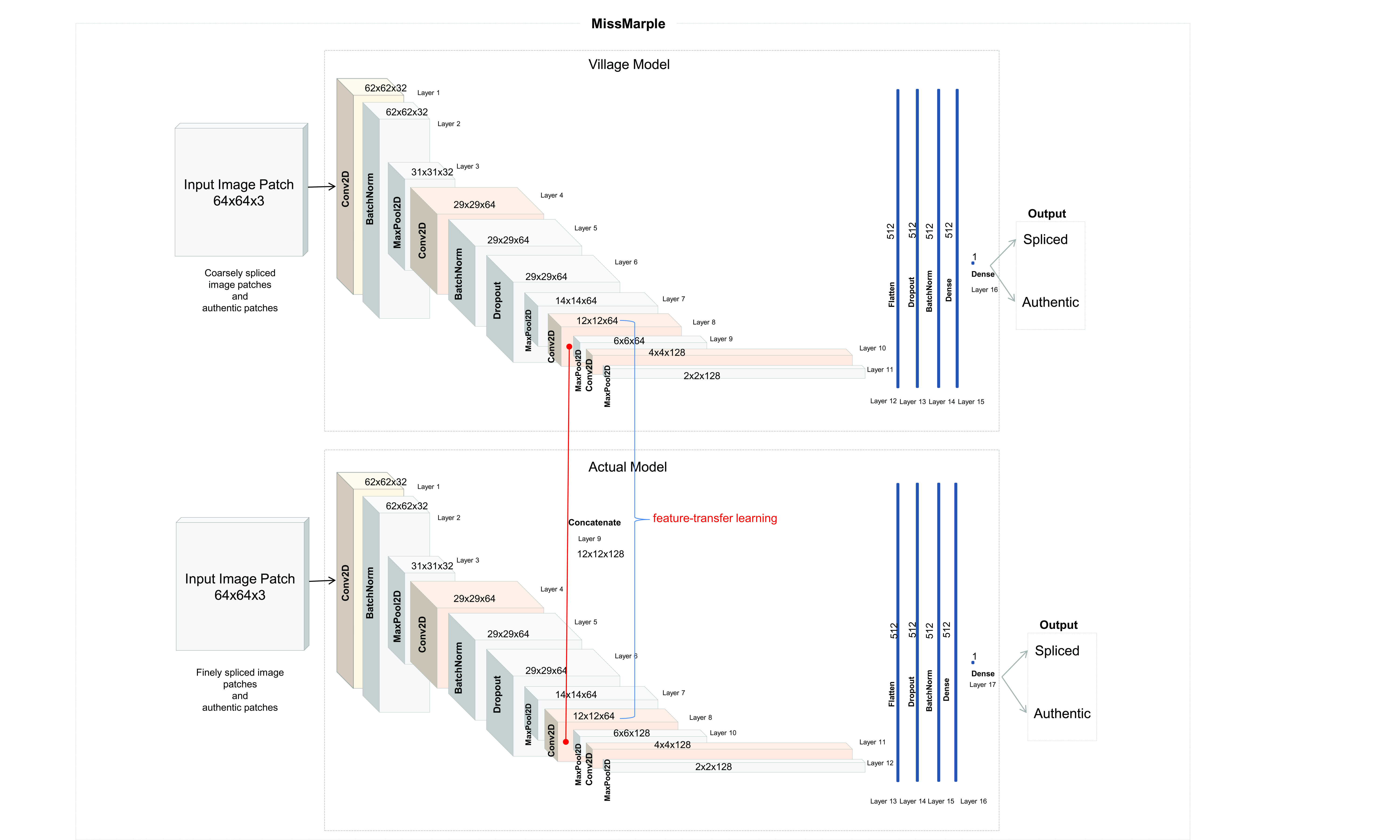}
	\caption{MissMarple proposed design}\label{fig:fig3}
\end{figure}

For the second part of the model, named the \textit{actual\_case\_model} (MM-A), we follow the same design layout as the MM-V model (twin network).
There is only one design change which is \textit{feature-transfer} learning. We first learn the features in the MM-V model on the coarsely spliced dataset C1 (refer to Table \ref{table:table2} for details on the dataset). Since the dataset has coarsely spliced forgeries, the boundaries surrounding the spliced regions are evident. We presume that the third convolutional layer learns features in the boundary well and generalises better. This can be seen in Figure \ref{fig:fig2}, where we generate heatmaps to visualise activations from the third convolutional layer of the MM-V model (V\_conv\_2d\_3). Considering the characteristics of convolutional neural networks, we know that the initial layers of the network learn local features and generalise these features in the progressive layers i.e. they learn spatial hierarchies. Additionally, the features learned by the convolutional layers are translation invariant (\citealp*{lawrence1997, lecun1998, simard2003}; \citealp[pg.~123]{chollet2016}). The features thus learned in V\_conv2d\_3 are then transferred to the MM-A model by sharing the weights learned in this layer. This is done by loading trained weights from V\_conv2d\_3 and operating them on the output of the second convolutional layer (A\_conv2d\_2) of the MM-A model. These weights are not trained during the training of the MM-A model. Finally, we concatenate the weights learned from V\_conv2d\_3 and A\_conv2d\_3 and input them to the last convolutional layer in the MM-A model.In totality we refer to this model as MM-V-A. Input to the second part are similar 64x64 fake (with 12.5\% overlap with the mask) and authentic patches except that we use a more realistic image splicing dataset. Equation (\ref{eq:eq1}) illustrates the feature-transfer learning followed in MM-V-A. Table \ref{table:table1} shows the hyperparameters chosen for the experiment while Figure \ref{fig:fig3} provides a detail layout of the proposed model design. Further details of the experimentation environment are discussed in section \ref{section:experimental_settings}.

The feature-transfer learning at the third convolutional layer in the \textit{actual\_case} model of the twin network can be explained with the help of the following equation. We adapt the representation from \citep{rao2017}. Let $F^n(X_{m})$ denote the feature map in layer $n$ of the convolution with kernel $W^n_m$ and bias $B^n_m$ for the input $X$ belonging to the $m$ model of the twin network. The non-linear activation function that is applied to each element of the input data is denoted as $f^n_m$. Equation (\ref{eq:eq1}) represents the feature-transfer learning.

\begin{equation}
F^3(X_A) = (F^2(X_A) * W^3_{V_{trained}})~concat~ (f^3_A(F^2(X_A) * W^3_A + B^3_A)) \label{eq:eq1}
\end{equation}

\begin{table}
	\tbl{Choice for hyperparameters} 
	{\begin{tabular}{l l} 
		\toprule
		Hyperparameter & Value \\ 
		\midrule
		Total iterations & 100 \\
		No. of epochs in each iteration & 30 (employed keras.callbacks for early stopping)\\
		Optimizer & RMSprop \\
		Learning rate & 1e-4 \\
		Activation in the dense layers & sigmoid \\
		Dropout rate & 0.1 and 0.5 \\
		\bottomrule
	\end{tabular}}
\label{table:table1}
\end{table}

Patch-based approach previously employed by Rao and Ni \citeyearpar{rao2017}, Mayer and Stamm \citeyearpar{mayer2018-unified} proved to be successful in detecting traces of image splicing. Another notable fact is that unlike previous CNN  architectures (\citealp{rao2017, bayar2016, cozzolino2019, rao2020}; Yaqi Liu \& X. Zhao \citeyear{liu2020}) that constrain the earlier layers of their model to suppress image content, our proposed MM-V-A is free from that restriction. Thus \textit{MissMarple} is a socio-inspired unconstrained twin CNN model with feature-transfer learning for image splicing detection.

\section{Experimentation Environment}\label{section:experimental_settings}
\subsection{Datasets.} 
\begin{figure}
	\centering
	\includegraphics[width=10cm, keepaspectratio]{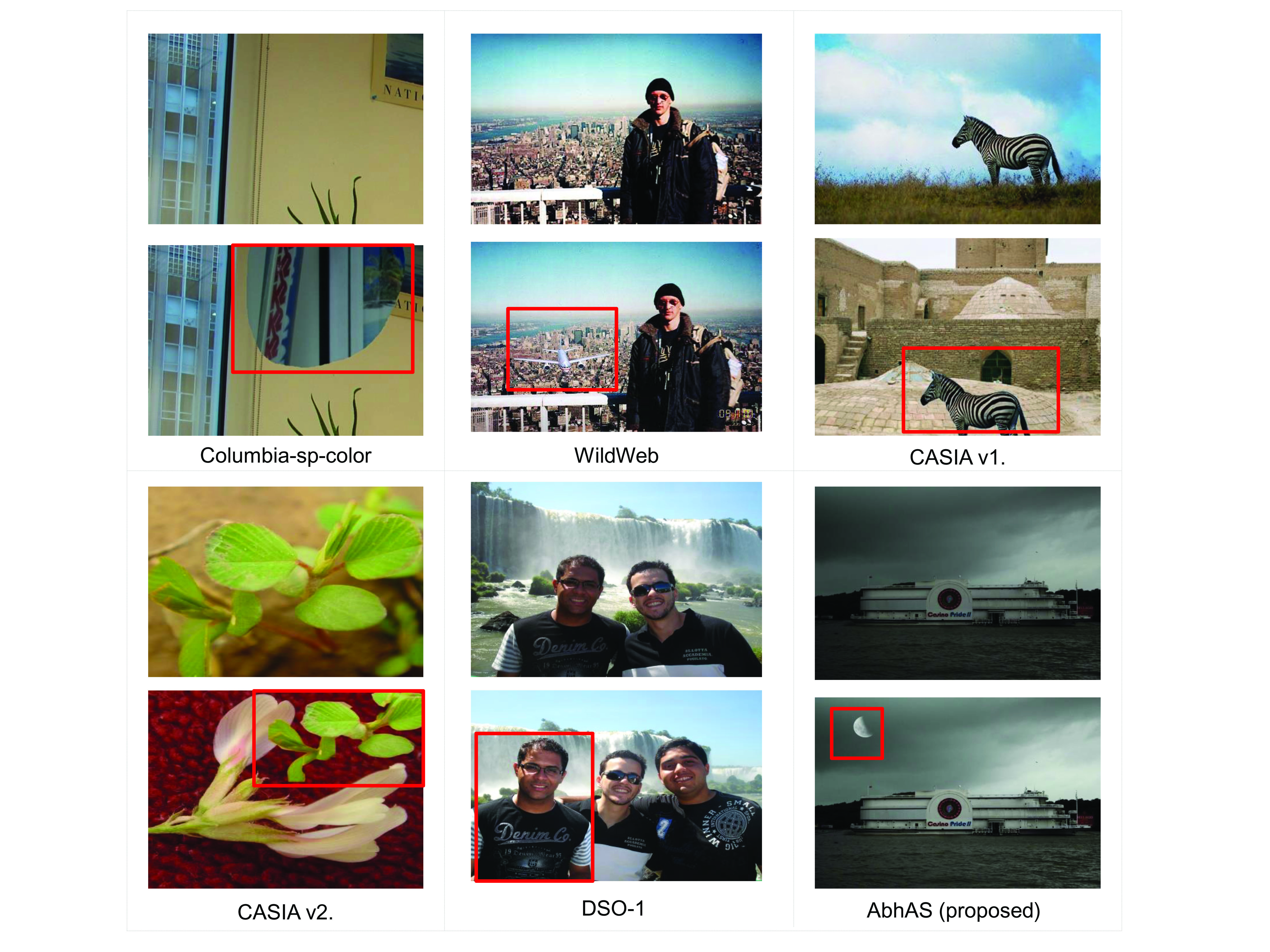}
	\caption{Sample authentic and spliced images from Columbia (Hsu \& S.F. Chang, \citeyear{hsu2006}), WildWeb \citep{zampoglou2015}, CASIA v1 and v2 \citep{dong2013-casia}, DSO-1 \citep{carvalho2013} and the proposed AbhAS datasets\citep{Gokhale2020abhas} where the spliced regions have been manually marked with a red bounding box} \label{fig:fig4}	
\end{figure}

We trained our \textit{MissMarple-village model} using the Columbia color image splicing dataset (Columbia-sp-color) (Hsu \& S.F. Chang, \citeyear{hsu2006}) which consists of 180 spliced and 183 authentic images. For the \textit{MissMarple-actual case model}, we train it using the DSO-1 dataset \citep{carvalho2013} consisting of 100 spliced and 100 authentic images. To further verify this claim, we also use the Wild Web tampered image splicing dataset \citep{zampoglou2015} consisting of over 9666 image splices obtained from 82 unique cases of forgery and their respective unaltered images. We selectively shortlisted 198 spliced images, with each image splice unique in nature. The selection also depended on the shape of the authentic image, a necessary check to manually generate additional masks (not part of the original dataset) for the patch based approach (refer to section \ref{section:design}). The number of authentic images corresponding to each splice were 99 in all. We may not follow the definition of authentic images for the Wild Web dataset because all the images were acquired from the Internet. The DSO-1 dataset on the other hand includes authentic images true to its definition of acquiring using single source camera. We follow a patch based approach to image splicing detection (refer to section \ref{section:design}). Details of the number of training, validation and testing samples (image patches) can be found in Table \ref{table:table2}. Figure \ref{fig:fig4} shows sample images from that dataset.

\begin{table}
	\tbl{Details of the Datasets Used}	
	{\begin{tabular}{l l l l l l l l}
		\toprule
		Code & Dataset & Spliced & Authentic & Total & Training & Validation & Test \\ 
		& & patches & patches & patches & patches & patches & images \\
		\midrule
		C1 & Columbia-sp-color & 1698 & 1728 & 3426 & 2398 & 1028 &   73 \\
		F1 & DSO1 &  4331  &  4320 & 8651  & 6055  & 2596  &  40 \\
		F2 & WildWeb  &   {4079}   &  {4053}   &  8132   & 5007 & 2147 &   59 \\
		F5 & AbhAS (proposed) &   5528 &   5523 & 11051  & 7735  & 3316  &   19 \\
		\bottomrule
	\end{tabular}}
\label{table:table2}
\end{table}

\subsection{Implementation}We implemented the code in Python using the deep learning framework \textit{Keras} \footnote{https://keras.io/} which is part of the \textit{Tensorflow} \footnote{https://www.tensorflow.org/} library and executed it in the \textit{Kaggle} environment \footnote{https://www.kaggle.com/}, availing the GPU facility.

\subsection{Experimental Trials}
We present to you the nature of experimental trials conducted using the \textit{MissMarple} socio-inspired convolutional neural network. Initially, we conducted a series of tests to finalise the values for hyperparameters. Keeping those constant (Table \ref{table:table1}), we proceed to justify the importance of feature-transfer learning \citep{Pan10} (Section \ref{section:transfer_learning}). 
\begin{enumerate}
	\item Firstly, we train the \textit{MissMarple-actual case model}(MM-V) alone using the fine splicing dataset F1 to nullify any doubts regarding the effect feature-transfer learning can have on the MM-A model.
	\item Secondly, we justify our claim that feature-transfer learning from a model trained on a coarsely spliced dataset to a model trained on the finely spliced dataset has a positive effect on image splicing detection. To do this, we first train the \textit{MissMarple-village model} (MM-V) using the coarsely spliced dataset C1 and share weights from the \textit{V\_conv2d\_3} with the \textit{A\_conv2d\_3} layer of the MM-A network which is trained using dataset F1. Initial empirical results showed that the third convolutional layer provides optimal feature-transfer learning. Further, the experimental trials are going forward with this design.
	\item Next, in order to validate the significance of result improvement with feature-transfer learning discussed in point 2 we use two additional finely spliced datasets F2 and F3 to train our MM-V model and repeat the experiment.
\end{enumerate}

\subsection{Training and validation}We trained and validated our proposed model MM-V over 100 iterations over 30 epochs. We used the \texttt{keras.callbacks} library for early stopping in each epoch in order to avoid overfitting the model. The iteration for which we observed the highest validation accuracy was selected for feature-transfer learning in the MM-A part of the model. Again, the overall model MM-V-A was trained and validated for 100 iterations over 30 epochs with early stopping. The test results presented for MM-V and MM-V-A are using the model weights for the highest observed iteration's validation accuracy. It is important to note that the testing process involves a passive blind approach as illustrated in Figure \ref{fig:fig5}.

\subsection{Multiplications involved}
Table \ref{table:table3} shows the number of multiplications involved in our proposed model vs. \citep{rao2017} and \citep{pomari2018}. We consider the multiplications computed only for the convolutional and maxpooling layers. For computing the number of multiplications required, we consider the parameters involved in the convolution process found in Equation (\ref{eq:conv_mul}).

\begin{equation}
N \times M \times D_k^2 \times D_p^2  \label{eq:conv_mul}
\end{equation}

Where, \\
$N$ = number of filters (or the output channels) of the convolutional layer.\\
$M$ = number of channels of the filter = number of channels of the input image.\\
$D_k^2$ = shape (rows and columns) of the kernel.\\
$D_p^2$ = shape (rows and columns) of the resultant output of that convolutional layer. 

Though the \citep{pomari2018} model consists of an averagepooling layer, we do not consider it during computation of the number of multiplications. Additionally, the \citep{pomari2018} model preprocesses the input images to obtain their illuminant maps. For this, the authors processed images belonging to the DSO1 dataset \citep{carvalho2013}. The average size of an image from the dataset is $2048\times1536$. In order to compare the results of multiplication and for the sake of simplicity, we consider the smallest size that the model can handle which is $256\times256$. The rest of the details can be found in Table \ref{table:table3}. From the table it is clear that the difference in the number of total mulitplications between the \textit{MissMarple} and \citep{rao2017} model is 24967098 which suggests that our proposed model is 68.3890\% faster than the \citep{rao2017} model. On the other hand, difference in the number of total multiplications between the \textit{MissMarple} and \citep{pomari2018} is 1285397376 which means that our proposed model is 99.1102\% faster than the Pomari et al. model. 

\begin{table}
	\tbl{Comparison on model design mutiplications} 
	{\begin{tabular}{l l l r} 
		\toprule
		Model & Input & No. of  & No. of total \\[-0.3em]
		& shape & convolutional layers & multiplications \\
		\midrule
		\citep{rao2017} & $128\times128\times3$ & 8 convolutional & 36507450\\
		& & 2 maxpooling & \\
		\citep{pomari2018} & $256\times256\times3$ & 52 convolutional & 1296937728\\
		& \tiny{(the smallest considered)} & 1 maxpooling & \\
		& & 1 averagepooling & \\
		& & \tiny{(not considered)}&  \\
		MissMarple & $64\times64\times3$ & \textit{MM-V} & 11540352\\
		(proposed) & & 4 convolutional & \\
		& & 4 maxpooling & \\
		& & \textit{MM-A} & \\
		& & 5 convolutional & \\
		& & 4 maxpooling & \\
		\bottomrule
	\end{tabular}}
\label{table:table3}
\end{table}

\subsection{Localisation} \label{section:localization}
\begin{figure}
	\centering
	\includegraphics[width=15cm, keepaspectratio]{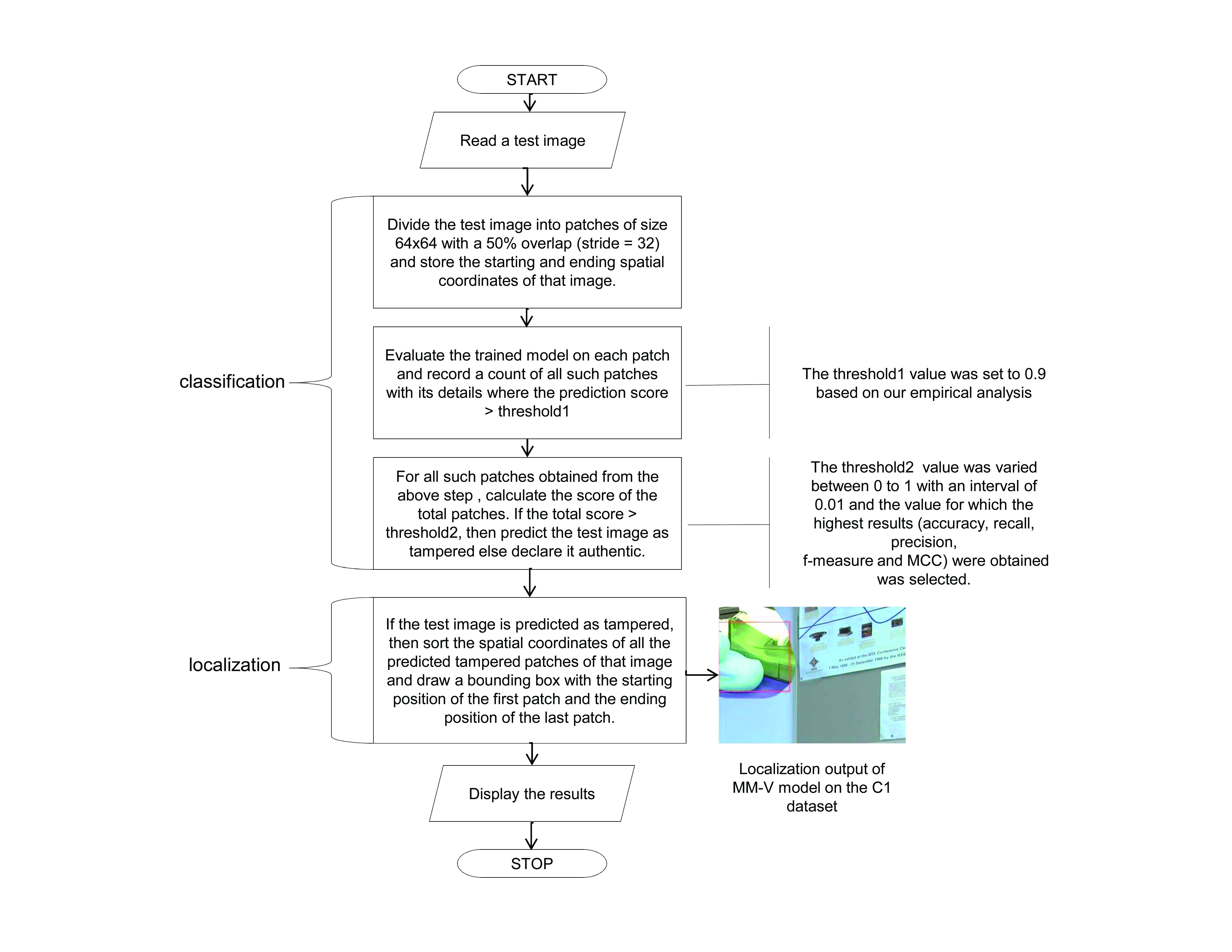}
	\caption{Flowchart depicting the passive blind image splicing classification and localisation procedure}
	\label{fig:fig5}
\end{figure}

We extended the proposed model to localise the spliced regions. Here, after the test image has been predicted as tampered, we employ a simple bounding box technique to outline the forged region. The flowchart in Figure \ref{fig:fig5} elaborates the passive blind splicing classification and localisation technique for one test image. The procedure is repeated for all the test images and results of classification are discussed in section \ref{section:results}. The flowchart also displays the result of localisation for a test image from the Columbia-sp-color (C1) dataset evaluated using the \textit{MissMarple} village model (MM-V). 

\subsection{Evaluation Metrics}
We briefly define the evaluation metrics presented in the paper. Considering that the proposed method selects patches from the fake and authentic regions as adapted from \citep{Sin19}, the dataset is balanced. The most common measure to compare classification performance of the model in case of balanced datasets is accuracy\footnote{also referred to as detection accuracy or classification accuracy in literature.}. Again, as our interest lies in correctly identifying fake images, we expect to reduce the false negative rate and hence select recall\footnote{also referred to as detection rate or true positive rate in literature.} as our second important measure. Finally, we present results obtained for precision, f-measure and MCC for our model performance. When comparing our proposed work with earlier studies, we limit to accuracy alone. The measures \footnote{Note: TP = True positives, TN = True negatives, P = Total spliced (positive) images and N = Total authentic (negative) images.} are defined in Table \ref{table:table4}. 

\begin{table}
	\tbl{Evaluation metrics} 
	{\begin{tabular}{l l}
		\toprule
		Measure	& Formula \\
		\midrule
		Accuracy & $\frac{TP + TN}{P + N}$ \\[0.3em]
		Recall & $\frac{TP}{P}$ \\[0.3em]
		Precision & $\frac{TP + FP}{P}$ \\[0.3em]
		F-measure & $2 \times \frac{Precision \times Recall}{Precision + Recall}$ \\[0.3em]
		Matthew's Correlation Coefficient (MCC) & $ \frac{(TP \times TN) - (FP \times FN)}{(TP + FP) \times (TP + FN) \times (TN + FP) \times (TN + FN)}$\\[0.3em]
		\bottomrule
	\end{tabular}}
\label{table:table4}
\end{table}

\section{Results and Discussion} \label{section:results}

\subsection{MissMarple model variants}
Table \ref{table:table5} displays the average results of the experimental trials over 100 iterations. The total time taken for training and validation is also reported rounded off to minutes and seconds. Table \ref{table:table6} provides details of the test results for the higest observed iteration's validation accuracy of model variants from Table \ref{table:table5}. Refer to Table \ref{table:table2} for the dataset code presented in the table.

\begin{table}[ht]
	\tbl{Average training and validation results of different experimental trials for 100 iterations} 
	{\begin{tabular}{l l l l l l l}
		\toprule
		Model  & Dataset & Train  & Train & Val     & Val & Total time taken  \\
		&		 & acc &  loss     & acc       & loss       &  (train + val)    \\
		\midrule
		MM-V   & C1     & 0.9873	& 0.0349   & \textbf{0.9387} & 0.2329     & 2m 58s \\
		MM-V   & F1     & 0.9359	&  0.1581  & \textbf{0.8338} & 0.5890  &  7m 11s \\
		MM-V-A & F1  & 0.9406	& 0.1467   &  0.8288         & 0.6338	  & 7m 14s \\	
		MM-V   & F2     & 0.9099	& 0.2101   & \textbf{0.7981} & 0.6637	  & 6m 59s \\
		MM-V-A & F2  & 0.9283	& 0.1716   &  0.7951         & 0.829	  & 7m 12s \\
		\bottomrule	
	\end{tabular}}
\tabnote{\textsuperscript{a}Values in bold indicate the highest obtained validation accuracy for the respective dataset.}
\label{table:table5}
\end{table}

\begin{table}[ht]
	\caption{Test results for the highest observed iteration's validation accuracy of model variants from Table \ref{table:table5} (Note: T = threshold)}	
	{\begin{tabular}{l l l l l l l l l l}
		\toprule
		Model     & Dataset  & Iter & T & Acc & Recall & Prec & F1 & MCC  & Time \\ 
		&          &       &       &     &        &      & score   
		&       &  \\
		\midrule
		MM-V      & C1       & 67       & 0.02      & \textbf{0.9178}   & \textbf{0.9722}      & \textbf{0.8750}      & \textbf{0.9210}   & \textbf{0.8409}      & 2m 7s\\
		MM-V      & F1       & 74      &    0.04   &   \textbf{0.7750} & 0.8500 &  \textbf{0.7391}   &  \textbf{0.7907}   &   \textbf{0.5563} &  5m 40s \\
		MM-V-A    & F1       & 95        & 0.04      & 0.7500   & 
		\textbf{0.9000}     & 0.6923       & 0.7826   & 0.5241     & 5m 14s \\	
		MM-V      & F2       & 96        & 0.05      & 0.7458   & 
		0.9750               & 0.7358    & 0.8387    & 0.3682   & 0m 47s\\
		MM-V-A    & F2       & 85        & 0.03      & \textbf{0.7627}  & 
		\textbf{1.000}          & \textbf{0.7407}       & \textbf{0.8510}   & \textbf{0.4415}      & 0m 47s\\	
		\bottomrule	
	\end{tabular}}
\tabnote{\textsuperscript{a}Values in bold indicate the highest obtained evaluation metrics for the respective dataset.}
\label{table:table6}
\end{table}

From Tables \ref{table:table5} and \ref{table:table6}, we find that the village model, MM-V provides consistent results throughout the training accuracy (98.73\%), validation accuracy (93.87\%), testing accuracy (91.78\%) and the detection accuracy or recall (97.22\%) for the Columbia splicing dataset C1. The proposed feature-transfer learning model MM-V-A shows improvement in learning properties from a coarsely spliced dataset (C1) and transferring that learning to the finely spliced datasets (F1 and F2). This is also evident from the results obtained by testing the MM-V model alone on datasets F1 and F2. We find that the test accuracy for the MM-V-A model trained on the F2 dataset (76.27\%) shows improvement compared to the MM-V model trained on the F2 dataset, indicating the benefits of feature-transfer learning. However, the same is not true in the case of F1 dataset. Here, the MM-V model provides a comparatively higher test accuracy (77.50\%) compared to the MM-V-A model but we see an improvement in the recall of MM-V-A (90.00\%) over the MM-V model. This is not a discouraging factor since the MM-V-A model still performs better compared to existing work presented in the following section.

\subsection{Comparison with existing models}
Further, we compare the performance results from earlier work with the \textit{MissMarple} variants MM-V and MM-V-A on the different datasets. Tables \ref{table:table7}, \ref{table:table8} and \ref{table:table9} present a comparison of the detection accuracy obtained on different datasets. The column "Category", suggests different approaches to splicing detection. Block-based and key-point based methods, statistical characteristics, camera characteristics like color filter array (CFA), chromatic aberration, noise inconsistencies, compression artefacts, physical and geometrical properties explored for detecting traces of image splicing are included under the category \textit{digital signal processing} abbreviated as "DSP". Techniques involving statistical and machine learning classifiers to classify features engineered from DSP technique are recorded under the category \textit{machine learning} abbreviated as "ML and DSP" while those involving neural networks and fuzzy inference end-to-end models abbreviated simply as "ML". Similarly, \textit{deep learning} models used in combination with machine learning classifiers are abbreviated as "DL and ML" while those involving end-to-end deep learning models are abbreviated as "DL". For further details on categories refer to \citep[pg.~11, Table~6]{Gok20}. 

\begin{table}
	\tbl{Classification results on the Columbia-sp-color (C1) dataset}
	{\begin{tabular}{l l l}
		\toprule
		Model     & Category & Average Accuracy \\
		\midrule
		(Hsu \& S.F. Chang, \citeyear{hsu2006}) & DSP and ML & 90.74\\
		\citep{shi-chen-2007} & DSP and ML & 91.87 \\
		(Z. He et al., \citeyear{he-sun-2011}) & DSP and ML & 93.55 \\
		\citep{moghaddasi2014} &  DSP & 93.80 \\
		\citep{muhammad2014} & DSP & 96.39 \\
		\citep{hussain2013} & DSP & 94.17 \\ 
		\citep{alahmadi2017} & DSP & 97.77 \\
		(J.G. Han et al., \citeyear{han2016}) & ML & 92.89 \\
		(Q. Zhang et al., \citeyear{zhang-lu-2016}) & DSP & 94.10 \\
		(B. Chen et al., \citeyear{chen-qi-2017}) & DSP and ML & 100.00 \\
		\citep{pomari2018} & DL (SVM + RGB) & 89.00 \\
		& DL (SVM + IIC) & 81.00 \\
		& DL (SVM + GGE) & 77.00 \\
		MM-V (proposed) & DL & 93.87 \\
		\bottomrule
	\end{tabular}}
\label{table:table7}
\end{table} 

From Table \ref{table:table7} we see that the proposed model MM-V, a simple network of 4 convolutional layers (16 layer CNN) outperforms the existing deep learning model proposed by Pomari et al. \citeyearpar{pomari2018} with an average classification accuracy of 93.87\% over 100 iterations. Except the DSP models proposed by Muhammad et al. \citeyearpar{muhammad2014}, Hussain et al. \citeyearpar{hussain2013}, Alahmadi et al. \citeyearpar{alahmadi2017}, Q. Zhang et al. \citeyearpar{zhang-lu-2016} and B. Chen et al. \citeyearpar{chen-qi-2017}, MM-V shows considerable improvement over the other existing models either belonging to the "DSP" or the "DSP and ML" category. Hence, the premise on which the proposed model design began, where machine learning or deep learning models have not been able to provide a higher detection accuracy on coarsely spliced datasets \citep{zheng2019}, stands justified to a greater extent with the results obtained using MM-V. 

\begin{table}
	\tbl{Classification results on the DSO-1 (F1) dataset compared with earlier work.}
	{\begin{tabular}{l l l}
		\toprule
		Model & Category  & Average Accuracy \\
		\midrule
		\citep{pomari2018} & DL(SVM + RGB) & 61.00 \\
		& DL (SVM + IIC) &  96.00 \\
		& DL (SVM + GGE) &  69.00 \\
		MM-V-A & DL &  82.88 \\
		\bottomrule
	\end{tabular}}
\label{table:table8}
\end{table}

\begin{table}
	\tbl{Classification results on the WildWeb (F2) dataset compared with earlier work.}
	{\begin{tabular}{l l l}
		\toprule
		Model & Category & Average Accuracy \\
		\midrule
		(B. Chen et al., \citeyear{chen-qi-2017}) & DSP and ML &  100 \\
		MM-V-A & DL &  79.51 \\
		\bottomrule
	\end{tabular}}
\label{table:table9}
\end{table}

Tables \ref{table:table8} and \ref{table:table9} discuss the performance results obtained on the DSO-1 (F1) and WildWeb (F2) datasets compared to existing work by Pomari et al. \citeyearpar{pomari2018} and B. Chen et al. \citeyearpar{chen-qi-2017} respectively. Our proposed MM-V-A model provides a higher detection accuracy (average over 100 iterations) on the F1 dataset compared to two variants of \citep{pomari2018} model.

\subsection{Comparison with inductive self-taught transfer learning}\label{section:ctl}
We further demonstrate results of inductive self-taught transfer learning versus feature-transfer learning. In the case of inductive self-taught transfer learning \citep{Pan10} for splicing detection, $D_S \neq D_T$ and $T_S \neq T_T$. But the source and target domains as well as tasks though different are related. When the source domain labels are absent or not directly transferable (Raina, Battle, Lee, Packer \& A.Y. Ng, \citeyear{raina2007}), and the target labels are available, the source models learn by means of self-taught learning. For feature-transfer learning refer to section \ref{section:transfer_learning}. Tables \ref{table:table10} and \ref{table:table12} show the results of executing inductive self-taught transfer learning approach using pre-trained models Xception \citep{chollet2017} and ResNet50 (K. He et al., \citeyear{HeK16}) over 50 iterations. Here, we train only the final layer of the pre-trained models on the C1 and F2 datasets respectively. In the case of Xception model comprising a total of 132 layers, the training mode for all 129 layers except the last set of depthwise separable convolutional layers is set to false. In the case of ResNet50, with a total of 175 layers, we do not train the first 171 layers allowing only the final residual network to learn features from the images. For the Xception model, we extracted patches of size 72x72 since the minimum input shape required by the model architecture is 71x71. For the proposed \textit{MissMarple} model and ResNet50, we retain patches of size 64x64. Table \ref{table:table11} and \ref{table:table13} display the test results for the highest observed validation accuracy on C1 and F2 datasets respectively.

\begin{table}
	\tbl{Average training and validation results of inductive self-taught transfer learning vs. feature-transfer learning on the Columbia splicing (C1) dataset.} 
	{\begin{tabular}{l l l l l l}
		\toprule
		Model    & Train  & Train & Val     & Val & Total time taken      \\ 
		&		  acc &  loss     & acc      & loss       &   (train + val)    \\
		\midrule
		MM-V       & 0.9873	& 0.0349   & 0.9387 & 0.2329     & 2m 58s \\
		Xception      & 0.9935	&   0.0224 & 0.9272 &   0.6451   & 4m 29s \\
		ResNet50    & 	0.9978 &   0.0065 & 0.9495 &  0.8029    &  4m 19s \\
		\bottomrule	
	\end{tabular}}
\label{table:table10}
\end{table}

\begin{table}
\tbl{Test results for the highest observed iteration's validation accuracy of model variants from Table \ref{table:table10} (T = Threshold)}
{\begin{tabular}{l l l l l l l l l}
	\toprule
	Model     & Iter & T & Acc & Recall & Prec & F1 & MCC   & Total time \\ 
	&        &       &     &        &      & score   
	&       & for testing \\ 
	\midrule
	MM-V          & 67        & 0.02      & 0.9178  & 0.9722      & 0.8750   & 0.9210   & 0.8409      & 2m 7s\\
	Xception      & 28 & 0.1 & 0.8767 & 0.9444 & 0.8293 & 0.8831 & 0.7610  & 7m 1s \\
	ResNet50    & 32 & 0.07 & 0.8904 & 1.000 & 0.8182 & 0.9000 & 0.8008 & 11m 0s\\
	\bottomrule	
\end{tabular}}
\label{table:table11}
\end{table}

\begin{table}
	\tbl{Average training and validation results of inductive self-taught transfer learning vs. feature-transfer learning on the WildWeb (F2) dataset.} 
	{\begin{tabular}{l l l l l l}
		\toprule
		Model   & Train  & Train & Val     & Val & Total time taken      \\ 
		&		  acc &  loss     & acc      & loss       &   (train + val)    \\
		\midrule
		MM-V-A   & 0.9283	& 0.1716   &  0.7951         & 0.829	  & 7m 12s \\
		Xception   & 0.9783	&  0.0613  &  0.8400        & 1.1147	  & 7m 27s\\
		ResNet50   & 	0.9940 &  0.0169  & 0.7420   &  2.1001  & 7m 2s \\
		\bottomrule		
	\end{tabular}}
\label{table:table12}
\end{table}

\begin{table}
\tbl{Test results for the highest observed iteration's validation accuracy of model variants from Table \ref{table:table12} (Note: T = threshold)}
{\begin{tabular}{l l l l l l l l l}
	\toprule
	Model      & Iter & T & Acc & Recall & Prec & F1 & MCC       & Total time \\ 
	&       &       &     &        &      & score   
	&       & for testing \\ 
	\midrule
	MM-V-A       & 85        & 0.03      & 0.7627   & 
	1.000          & 0.7407       & 0.8510   & 0.4415     & 0m 47s\\	
	Xception    & 37 & 0.05 & 0.7458 & 1.000 & 0.7273 & 0.8421 & 0.3913 & 2m 34s\\
	ResNet50   & 12 & 0.06 & 0.7119 & 1.000 & 0.7018 & 0.8247 & 0.2718 &  4m 41s\\	
	\bottomrule	
\end{tabular}}
\label{table:table13}
\end{table}

\subsection{Performance results on a proposed realistic splicing dataset}
\indent Additionally, we developed a new realistic splicing dataset for evaluation. The proposed dataset is referred to as "AbhAS" dataset with 45 authentic and 48 spliced images \citep{Gokhale2020abhas}. Some of these authentic images are original sources taken from the Internet while the remaining are true to the definition of authentic being captured from a single source camera (also refer section \ref{section:experimental_settings} for note on \textit{Datasets}). We reason that real-world splicing or image composite forgeries are created by downloading images from the Internet \citep{zampoglou2015}. In order to train our proposed model, we extract patches of size 64x64 from fake images and authentic images. For fake regions, we extract patches by overlaying the masks on fake images and select patches with an overlap of 30\%. To highlight the potential strength of this realistic dataset, we compared its performance with pre-trained models Xception and ResNet50. We followed a similar approach of training the pre-trained models as explained in section \ref{section:ctl}. Tables \ref{table:table14} and \ref{table:table15} present performance results on the AbhAS dataset (F3) thus encouraging researchers to further evaluate this dataset \citep{Gokhale2020abhas}. We observe from Table \ref{table:table14} that MM-V-A outperforms all other variants in terms of detection accuracy, precision, F1 score and MCC, indicating the potential of feature-transfer learning.

\begin{table}
	\tbl{Average training and validation results obtained on the AbhAS dataset \citep{Gokhale2020abhas} for 100 iterations} 
	{\begin{tabular}{l l l l l l}
		\toprule
		Model  & Train  & Train & Val    & Val & Total time taken      \\ 
		& acc &  loss     & acc       & loss       &  (train + val)    \\
		\midrule
		MM-V   & 0.9660 	& 0.0892   & 0.9374 &  0.2111    & 9m 44s \\
		MM-V-A &  0.9688	& 0.0831   & 0.9268 & 0.3104	  & 10m 36s \\	
		ResNet50 &  0.9925	& 0.0222  & 0.9504 & 0.2663	  &  14m 39s \\
		Xception &  0.9832	& 0.0508  & 0.9350 & 0.7337	  &  12m 3s \\
		\bottomrule		
	\end{tabular}}
\label{table:table14}
\end{table}

\begin{table}
\tbl{Test results for the highest observed iteration's validation accuracy of model variants from Table \ref{table:table14}. (Note: T = threshold)}	
{\begin{tabular}{l l l l l l l l l}
	\toprule
	Model     &  Iter & T & Acc & Recall & Prec & F1 & MCC       & Time \\ 
	&          &       &       &     &        &      &   
	&        \\
	\midrule
	MM-V        &   95     &   0.01    &  0.5263  &  0.9000   &   0.5294     &   0.6667 &   0.0181    & 7m 49s \\
	MM-V-A      & 97        & 0.02      & 0.6316   & 0.8000      & 0.6154       & 0.6957   & 0.2626     & 9m 18s \\
	ResNet50      & 17	 & 0.01      & 0.5789   &  0.9000  & 0.5625      & 0.6923    & 0.1674      & 45m 12s \\	
	Xception      & 45	 & 0.01      & 0.5263   & 0.9000  & 0.5294      & 0.6667    & 0.0181      & 28m 23s \\
	\hline	
\end{tabular}}
\label{table:table15}
\end{table}

\subsection{Summary of results and limitations}
Results obtained by training and testing MM-V on the Columbia image splicing dataset (Hsu \& S.F. Chang, \citeyear{hsu2006}) outperformed the deep learning model proposed by \citep{pomari2018} and some of the existing techniques belonging to digital signal processing, machine learning and a combination of these categories. The four convolutional neural network MM-V model also reported a higher test accuracy on the Columbia dataset (Hsu \& S.F. Chang, \citeyear{hsu2006}) compared to the popular pre-trained models namely, Xception \citep{chollet2017} and ResNet50 (K. He et al., \citeyear{HeK16}); indicating that an end-to-end deep learning model built using fewer layers can perform better than models built using an inductive self-taught transfer learning approach. The complete feature-transfer learning model, MM-V-A trained and tested on the DSO-1 dataset\citep{carvalho2013}, outperformed two variants proposed by Pomari et al.'s \citeyearpar{pomari2018} deep learning models. The performance of the MM-V-A model on the WildWeb dataset \citep{zampoglou2015} was higher compared to the pre-trained models Xception \citep{chollet2017} and ResNet50 (K. He et al., \citeyear{HeK16}); indicating that feature-transfer learning improves the detection accuracy over inductive self-taught transfer learning. Additionally, we also trained and tested the MM-V-A model on a proposed realistic image splicing dataset referred as AbhAS \citep{Gokhale2020abhas}, obtaining better results compared to inductive self-taught transfer learning approach using Xception \citep{chollet2017} and ResNet50 (K. He et al., \citeyear{HeK16}) pre-trained models.

\subsection{Testing on real datasets}
\begin{figure}
	\centering
	\includegraphics[width=5in, keepaspectratio, clip]{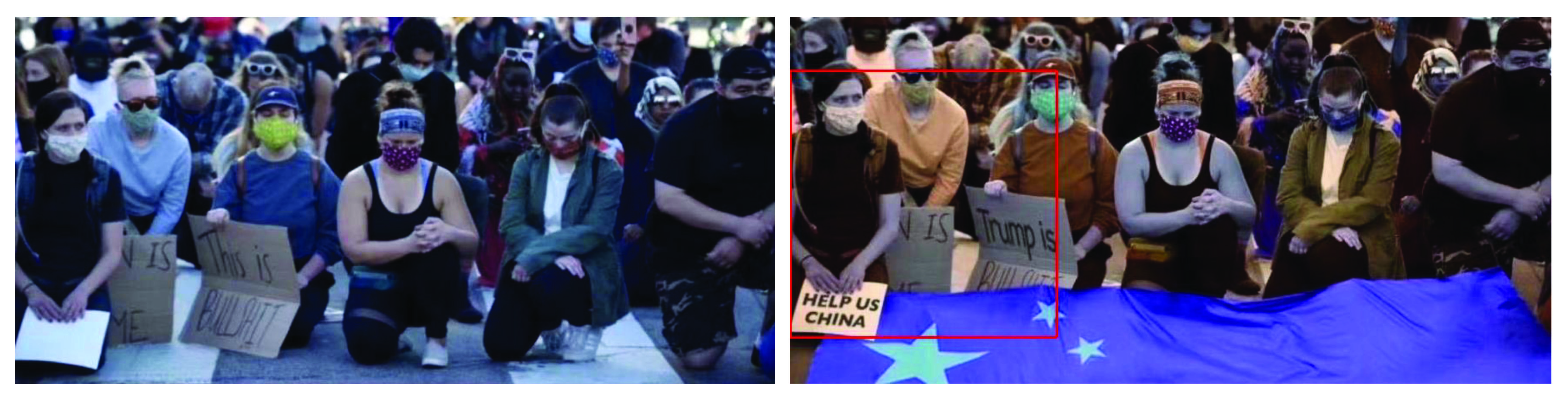}
	\caption{Sample authentic vs. spliced images obtained from reddit.com (refer footnote \ref{fn:fn1}) showing the result of our naive localisation \ref{fig:fig5}} \label{fig:fig6}
\end{figure}

In order to test the efficacy of our proposed model on real images obtained from social media, we downloaded 6 images (4 fake and 2 authentic) from the reddit website\footnote{https://www.reddit.com/r/photoshopbattles/comments/ay2twm/psbattle\_this\_cat\_wearing\_a\_suit/ by Hordon\_Gayward, Accessed July 2020}\footnote{https://www.reddit.com/r/China/comments/guknl9/fake\_images\_of\_us\_protests\_circulated\_on\_internet/ by Charlie\_Yu, Accessed July 2020}.\footnote{https://www.reddit.com/r/EnoughCommieSpam/comments/gv2p6e/another\_fake\_image\_of\_the\_us\_protests\_circulating/ by gucciAssVoid, Accessed July 2020 \label{fn:fn1}} \footnote{https://www.reddit.com/r/ExposurePorn/comments/hltotu/6144x3536\_3\_image\_composite\_of\_my\_buddies\_cars/ by Chucke412, Accessed July 2020}. We found that our MM-V-A models, trained on DSO1, WildWeb, proposed AbhAS and their ensemble could detect all the 4 spliced images correctly while the MM-V model could identify 3 out 4 images from the dataset. Our proposed naive localisation approach could mark the potential forged regions in 2 images (for all the MM-V-A models) while the MM-V-A model trained on the DSO1 dataset could also localise an additional image correctly. Figure \ref{fig:fig6} shows a sample localisation (refer footnote \ref{fn:fn1}).

\section{Conclusion}\label{section:discussion}
We presented a novel socio-inspired feature-transfer learning twin convolutional neural network (CNN) model, \textit{MissMarple} (MM-V-A), for image splicing detection. Features from the first part of the model referred to as \textit{village model} (MM-V) trained on a coarse spliced dataset, are transferred to the second part of the twin network referred to as \textit{actual-case model} (MM-A) to be trained on a finely spliced dataset. The proposed model consists of 9 convolutional layers and a total of 23 layers. Results obtained by training and testing MM-V on the coarsely spliced Columbia image splicing dataset (Hsu \& S.F. Chang, \citeyear{hsu2006}) and MM-V-A on the finely spliced DSO-1 dataset \citep{carvalho2013}, outperformed existing deep learning models and pre-trained models namely, Xception \citep{chollet2017} and ResNet50 (K. He et al., \citeyear{HeK16}); indicating that an end-to-end deep learning model built using fewer layers can perform better than models built using an inductive self-taught transfer learning approach. Additionally, we also trained and tested the MM-V-A model on a proposed realistic image splicing dataset referred as AbhAS \citep{Gokhale2020abhas}, and obtained better results compared to inductive self-taught transfer learning approach using pre-trained models. Though not all results obtained by the proposed \textit{MissMarple} model were satisfactory, we reason that our socio-inspired deep CNN architecture introduces a new line of thought demonstrating positive outcomes for a feature-transfer learning approach. We also demonstrated the potential of our model on fake news images obtained from the reddit.com website. Our future work requires to address improvements in the detection accuracy of our proposed \textit{MissMarple} model on the WildWeb \citep{zampoglou2015} and DSO-1 \citep{carvalho2013} datsets. The model also did not perform well on the CASIA TIDE datasets \citep{dong2013-casia} and we wish to employ a constrained image splicing detection approach. The current selection and finalisation of hyperparameteres for the \textit{MissMarple} model is based on empirical analysis and literature review. However, we shall focus on finetuning these parameters using standard built-in libraries in Keras. Finally, we invite researchers to conduct experiments on the proposed AbhAS dataset \citep{Gokhale2020abhas}. 

\section*{Acknowledgement(s)}
Angelina Gokhale would like to acknowledge Symbiosis International (Deemed University) for providing the junior research fellowship and Nikhil Aarons for extending help in creating the proposed realistic splicing dataset (referred as AbhAS). The authors wish to thank the Kaggle team for their coding platform, Fran\c{c}ois Chollet and the Google team for the Keras deep learning framework. The authors also express their sincere gratitude towards Mandaar Pande and Anil Jadhav for reviewing sections of this work and for their valuable and timely inputs.

\section*{Disclosure statement}
The authors declare that there are no competing interests.





\bibliographystyle{apacite}
\bibliography{mybibfile-MM}

\end{document}